\documentclass[runningheads]{llncs}
\usepackage{eccv}
\usepackage{eccvabbrv}
\usepackage{amsmath,amssymb}
\usepackage{pifont}
\usepackage{dsfont}
\usepackage{graphicx}
\usepackage{booktabs}
\usepackage{multirow}
\usepackage[accsupp]{axessibility}
\usepackage{hyperref}
\usepackage{orcidlink}
\makeatletter
\newcommand{\printfnsymbol}[1]{%
  \textsuperscript{\@fnsymbol{#1}}%
}
\makeatother

\begin{document}

\title{DiffClass: Diffusion-Based Class Incremental Learning}

\author{Zichong Meng\inst{1} \and
Jie Zhang\inst{2} \and
Changdi Yang\inst{1} \and
Zheng Zhan\inst{1} \and
Pu Zhao\thanks{Corresponding Author}\inst{1} \and 
Yanzhi Wang\printfnsymbol{1}\inst{1}}

\authorrunning{Z.~Meng et al.}
\institute{Northeastern University, Boston MA 02115, USA \and
ETH Zürich, 8092 Zürich, Switzerland}

\maketitle

\begin{abstract}
Class Incremental Learning (CIL) is challenging due to catastrophic forgetting. On top of that, exemplar-free CIL is even more challenging due to forbidden access to data of previous tasks. Recent exemplar-free CIL methods attempt to mitigate catastrophic forgetting by synthesizing previous task data. However, they fail to overcome the catastrophic forgetting due to the inability to deal with the significant domain gap between real and synthetic data.
To overcome these issues, we propose a novel exemplar-free CIL method.
Our method adopts multi-distribution matching (MDM) diffusion models to align quality of synthetic data and bridge domain gaps among all domains of training data. Moreover, our approach integrates selective synthetic image augmentation (SSIA) to expand the distribution of the training data, thereby improving the model's plasticity and reinforcing the performance of our multi-domain adaptation (MDA) technique. With the proposed integrations, our method then reformulates exemplar-free CIL into a multi-domain adaptation problem to implicitly address the domain gap problem and enhance model stability during incremental training.
Extensive experiments on benchmark CIL datasets and settings demonstrate that our method excels previous exemplar-free CIL methods with non-marginal improvements and achieves state-of-the-art performance. Our project page is available at \url{https://cr8br0ze.github.io/DiffClass}.
\end{abstract}

\section{Introduction}
\label{sec:introduction}
Although recent deep learning (DL) models have achieved superior performance even better than humans in various tasks, catastrophic forgetting~\cite{catastrophic_forgetting} remains a challenging problem that limits the continual learning capabilities of DL models. Unlike humans, DL models are unable to learn multiple tasks sequentially, which forget the previous learned knowledge after learning new tasks. To address this, Class Incremental Learning (CIL) extensively investigates how to learn the information of new classes without forgetting past knowledge of previous classes.  Various CIL works~\cite{icarl,ucir,podnet,foster, beef} try to untangle catastrophic forgetting through saving a small proportion of previous training data as exemplars in memory and retraining with them in new tasks.  However, these methods suffer from privacy and legality issues of utilizing past training data, as well as memory constraints on devices. Different from previous exemplar-based CIL, Exemplar-Free CIL~\cite{abd, rdfcil, ddgr} has gained increasing popularity where DL models incrementally learn new knowledge without storing previous data as exemplars.

To counteract forgetting knowledge of past tasks, the most recent exemplar-free CIL works~\cite{abd, rdfcil, ddgr, zhang2023target} propose to synthesize previous data instead of using real data.  The synthetic data of previous tasks are generated through either model inversion~\cite{model_inversion} with knowledge distillation or denoising diffusion models~\cite{diffusion}.
However, these methods suffer from significant domain gaps between synthetic data and real data especially when the number of incremental tasks is large (\ie long-term CIL), which inevitably misleads the decision boundaries between new and previous classes.  The obtained models favor plasticity over stability, meaning they tend to learn new knowledge but without keeping previous knowledge in mind as demonstrated in \cref{sec:domaingaps}. Therefore,  how to exhibit both stability and plasticity in exemplar-free CIL remains a crucial challenge. 

To address these problems, we propose a novel exemplar-free CIL approach that bridges the crucial domain gaps and balances stability and plasticity.
Our method incorporates a multi-distribution-matching (MDM) technique to finetune diffusion models resulting in closer distributions between not only synthetic and real data but also among synthetic data through all incremental training phases.
Our method also reformulates exemplar-free CIL as task-agnostic multi-domain adaptation (MDA) problems to further deal with domain gaps between real and synthetic data, with selective synthetic image augmentation (SSIA) to enhance each incremental task learning with current task synthetic data. 

We summarize our contributions as follows:
\begin{itemize}
    \item We introduce a novel exemplar-free CIL method that explicitly mitigates forgetting and balances stability \& plasticity by adopting MDM diffusion models and enhancing the dataset with SSIA
    to address domain gaps in exemplar-free CIL settings.
    \item We propose an innovative approach to reformulate exemplar-free CIL as task-agnostic MDA problems. This groundbreaking step implicitly manages domain gaps during CIL training, better addressing catastrophic forgetting in exemplar-free CIL.
    \item Extensive experiments on CIFAR100~\cite{cifar} and ImageNet100~\cite{imagenet}  demonstrate that our method effectively mitigates catastrophic forgetting in different exemplar-free CIL settings, surpassing SOTA methods with significant improvements.
\end{itemize}

\section{Related Work}
\label{sec:related_works}
\subsubsection{Class Incremental Learning (CIL)} has merged as a challenging problem focusing on how models can incrementally learn new classes without forgetting previously acquired knowledge. 
To overcome the catastrophic forgetting,
recent successful approaches~\cite{icarl,ucir,podnet,foster, beef,zhu2024model,zhu2023generalized,zhu2023universal,zhu2023bridging, sparcl, dualhsic} store training data from previously learned classes as exemplars and replay them while learning new tasks. Exemplars are indeed helpful for reviewing past task knowledge and thus benefit the incremental learning process. However, due to privacy and legality issues, and memory constraints on devices, it may be unachievable in practice. 

Thus, exemplar-free CIL gains increasing popularity among researchers. In recent years, instead of using exemplars, several exemplar-free CIL methods~\cite{lwf, related1, related2, relate3, relate4, relate5} synthesize images of previously learned classes as a review instead of storing real images to mitigate forgetting. However, most of these methods suffer from significant performance degradation due to large domain gaps between synthetic and real data. Later methods~\cite{abd, rdfcil} propose to utilize modified knowledge distillation techniques to constrain the domain gaps. These methods fail as knowledge distillation tends to attribute the suboptimal performance from the previous task to the model's learning capabilities in each current task, especially with the domain gaps in data. 
Different from previous works, our framework specifically aims
to bridge domain gaps in exemplar-free CIL leading to a model with robustness in both stability and plasticity.

\subsubsection{Diffusion model.} Diffusion models~\cite{diffusion, diffusion1, diffusion2} generate images through stochastic differential equations by progressively denoising them. This technique involves two primary stages: a forward diffusion process that incrementally introduces Gaussian noise into the input data, and a reverse diffusion process that is trained to gradually reverse this procedure, effectively removing the noise from noised input data.

Subsequent research has focused on enhancing the quality of generated outputs. There are  various methods, including scaling the model~\cite{diffusion3, diffusion5, diffusion6, diffusion, v1.5, zhang2023federated}, and refining the training and sampling processes  ~\cite{diffuion4, diffusion8, diffusion9}.
Among them, the Latent Diffusion Model (LDM) stands out for exhibiting great text-to-image generation quality due to scaling up the diffusion model by conducting both forward and backward diffusion processes in latent space.

In recent developments, diffusion models have also shown remarkable versatility beyond image generation. They are successfully applied in various domains, including other computer vision tasks~\cite{promptdiff, instructgie}, audio processing~\cite{audiodiff}, and even text-to-3D~\cite{3ddiff}.

How to customize efficiently customized diffusion models also stands out as a recent heated topic. Various works have been proposed including techniques involving altering text embeddings~\cite{texualinversion, aestheticgradient}, altering text embeddings~\cite{texualinversion, aestheticgradient}, associating special words with small number of example images~\cite{dreambooth}, or inserting a small number of new weights~\cite{lora},

In this work, we employ the Stable Diffusion model, \ie a variant of LDM, and tailor it specifically for our method using LoRA~\cite{lora} that can effectively address domain gaps in exemplar-free CIL.
\section{Diagnosis: Domain Gaps in Exemplar-Free CIL}
\label{sec:domaingaps}
\begin{figure}[tb]
\centering
\includegraphics[width=190pt, height=135pt]{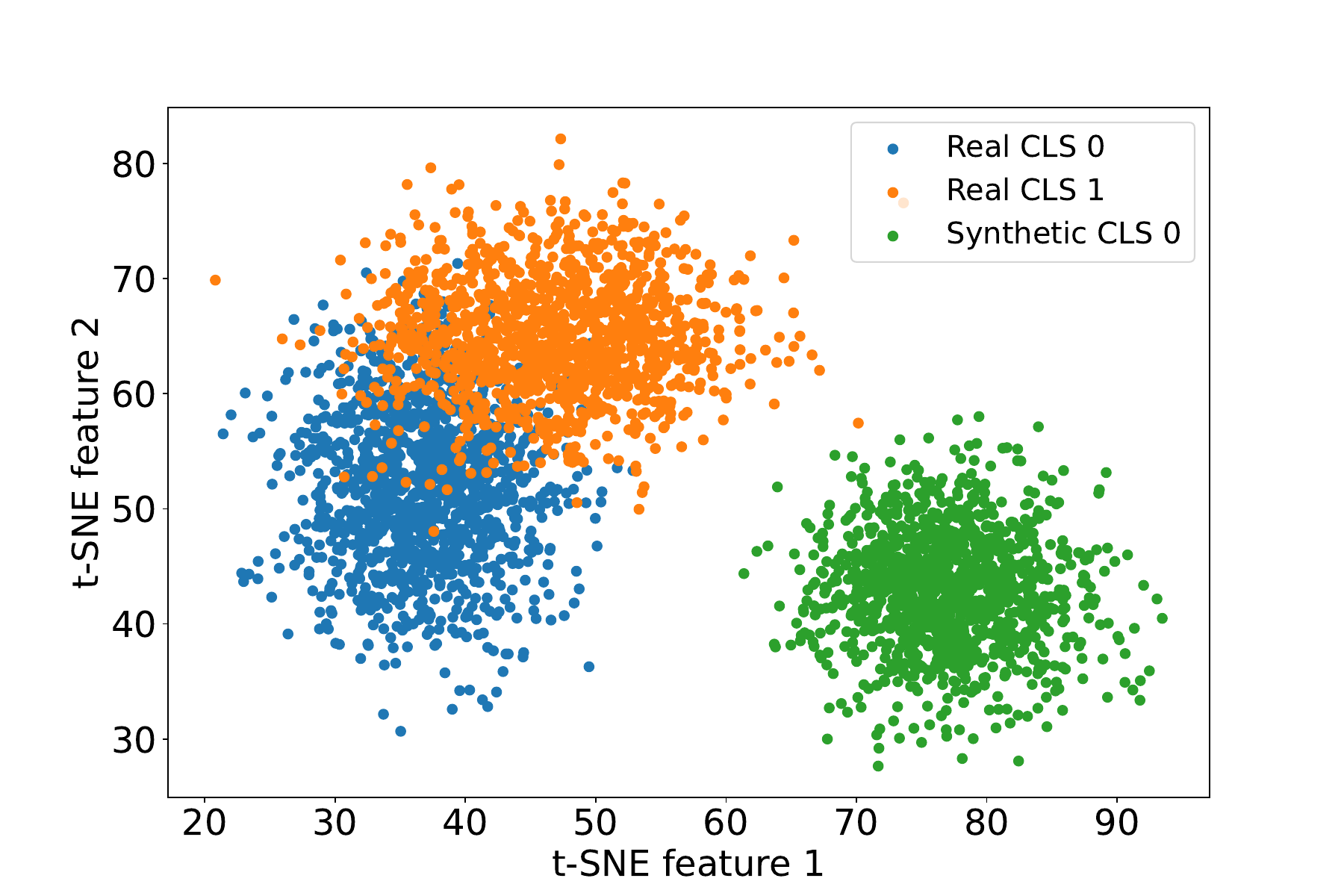}
\caption{\textbf{Domain Gaps in Exemplar-Free CIL}. The distribution of real classes is closer to each other while domain gaps exist between real class 0 and synthetic class 0.}
\label{fig:domain1}
\end{figure}
Although recent advancements in generative artificial intelligence can generate realistic images, we notice that the distributions of the generated synthetic images are still different from those of real images with domain gaps, leading to low accuracy in the classes trained with synthetic data in exemplar-free CIL settings.  We also further dig into the low accuracy and find that the reason may be the model's preference for domains over classes after training, \ie the model classifies whether the image is real or synthetic rather than its true label.     

In \cref{fig:domain1}, a t-SNE visualization is performed to compare real data of class 0 and 1 from ImageNet100~\cite{ucir} with synthetic data of class 0 generated by the pretrained stable diffusion V1.5 model~\cite{v1.5}. The visualization reveals that the distributions of the real classes are more closely aligned, while a significant domain gap is evident between the synthetic data of class 0 and its real counterpart.

These domain gaps can potentially effect model's performance after model training with real and synthetic data, since the decision boundary can be significantly distorted by synthetic data, as it may treat the real class 0 and class 1 (with a smaller distribution discrepancy) as the same class in testing.

We also conduct an experiment in a class incremental setting to further verify. In specific, we train a model with only a ResNet~\cite{resnet} backbone and a linear classifier for the first four tasks (each with 5 classes) in a 20-task CIL setting on the ImageNet100 dataset (refer to \cref{sec:experiment} for more details).
From the second to the fourth tasks, aside from the real data of the current task, we also train with synthetic data of the previous tasks generated by the pre-trained SD V1.5 model.  
We additionally train another model with entirely real data for the four tasks as a reference for how well the model can perform with real data.
{\setlength{\tabcolsep}{4.3pt}
\renewcommand\arraystretch{1.25}
\begin{table}[tb]
  \centering
  \caption{Diagnosis experiment accuracy result (in \%) of incremental training the model with synthetic previous task data and real data of current task \vs 
  training model with all real data for first four tasks of twenty-task incremental setting on ImageNet100. }
  \label{tab:domain}
  \begin{tabular}{l|c|c|c|c|c}
  \hline
   Training Data Domain & CLS 0-4 & CLS 5-9 & CLS 10-14 & CLS 15-19 & Total Classes \\
   \hline
   Synthetic \& Real & 47.67 & 48.39 & 51.11 & 89.31 & 59.37\\
   Real Data Only & 85.97 &  80.11 & 83.54 & 81.27 & 82.72\\
   \cline{2-4}
  \hline
  \end{tabular}
\end{table}
}

In \cref{tab:domain}, we present the accuracy on the real test dataset at the end of task 4. As observed, the model performs significantly better on the classes of the new task (\ie class 15-19, trained with real data) than previous tasks (\ie class 0-14, trained with both real in previous task and synthetic data in the current task), demonstrating the model's preference for plasticity over stability.

\begin{figure}[tb]
  \centering
  \begin{subfigure}{\linewidth}
    \centering
    \begin{subfigure}{0.495\linewidth}
      \includegraphics[width=\linewidth]{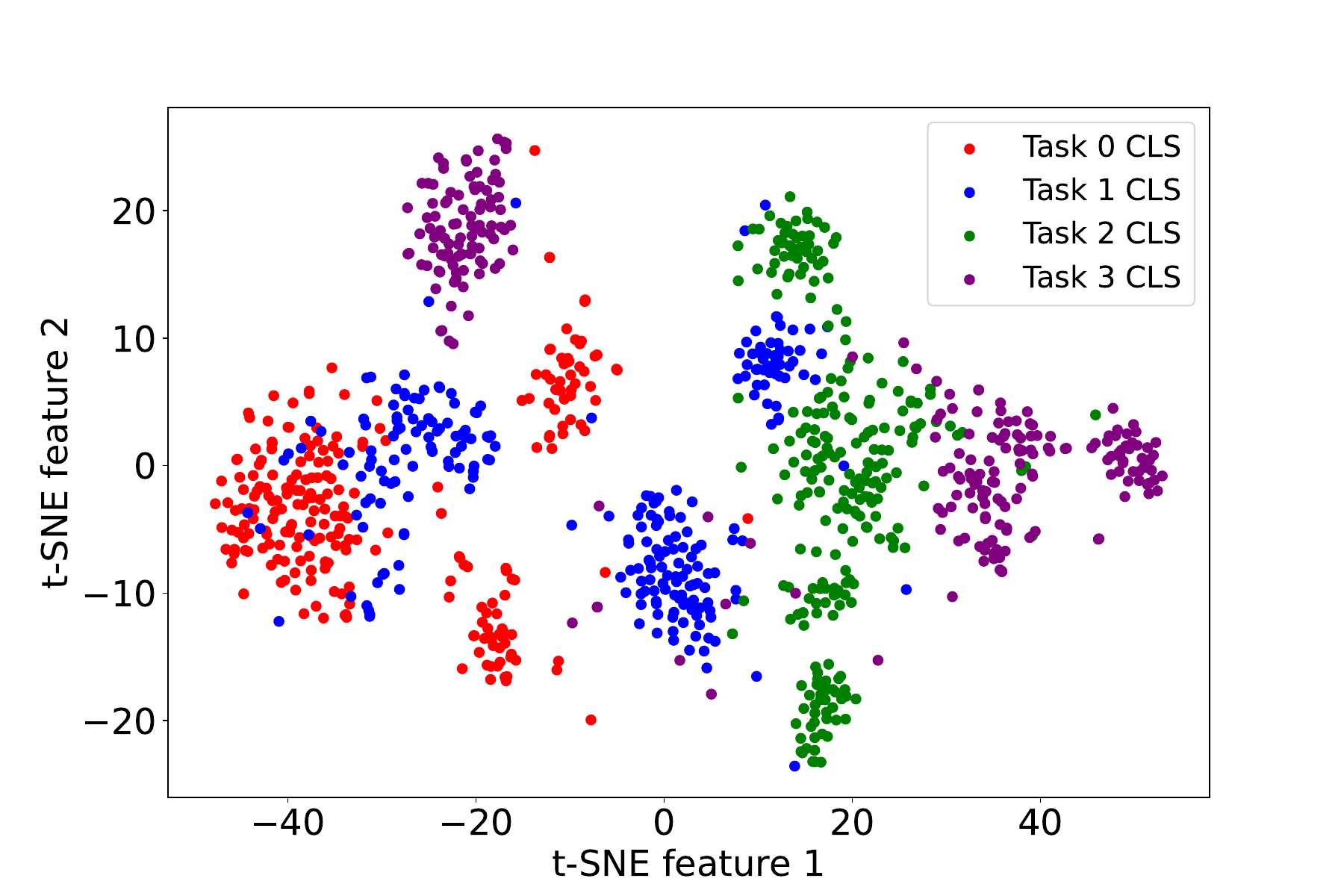}
      \caption{Feature Embedding with Ground Truth Label}
    \end{subfigure}
    \begin{subfigure}{0.495\linewidth}
      \includegraphics[width=\linewidth]{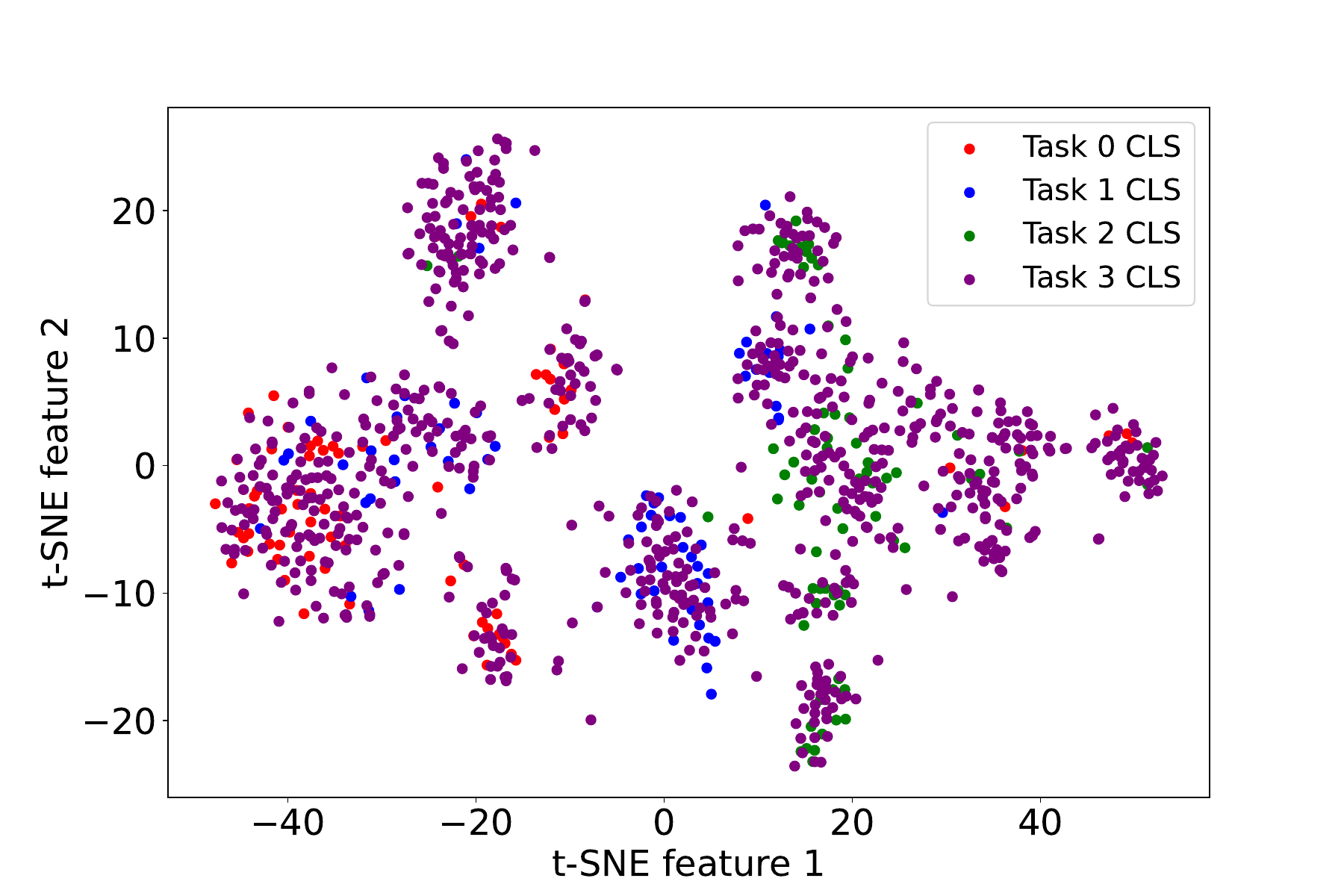}
      \caption{Feature Embedding with Predition Label}
    \end{subfigure}
  \end{subfigure}
  \caption{\textbf{t-SNE Visualization of Test Data's Feature Embedding.} Most of the previous task test data in incremental task 3 are misclassified as one of the task 3 classes.
  }
  \label{fig:domain2}
\end{figure}
In \cref{fig:domain2}, we further
use t-SNE visualization for the feature embeddings of test data extracted from the incrementally trained ResNet18 backbone. 
As observed from \cref{fig:domain2},
most of misclassified test data from classes of previous tasks are labeled to new classes of the most recent task,  
indicating the model's labeling preference of domain over class, \ie the model labels whether it is real or synthetic, rather than its true class.   

Inspired by the diagnosis experiments, our method tries to mitigate the domain gaps and balance plasticity \& stability.
\section{Methodology}
\label{sec:methodology}
\subsection{Framework}
Following previous works~\cite{icarl, abd, rdfcil}, CIL contains $N$ incremental learning phases or tasks.
In the $i^{th}$ incremental phase (or interchangeably $\mathcal{T}_{i}$) $0\leq i<N$, our framework mainly consists of the following three steps. 
\begin{figure}[t]
\centering
\includegraphics[width=\textwidth,keepaspectratio]{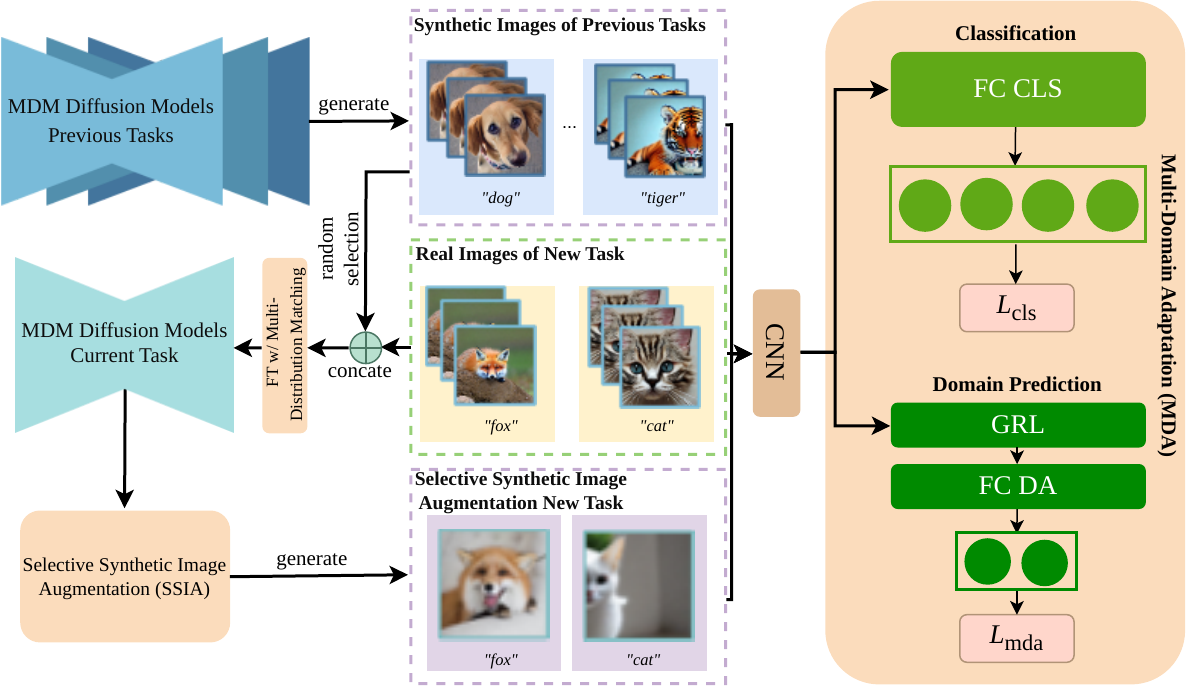}
\caption{\textbf{Model Framework Overview} learning on currect task $\mathcal{T}_{i+1}$. previous MDM diffusion models $J_{0:i}$ are used to generated Synthetic Data of previous tasks $\mathcal{D}_{0:i}^{\text{syn}}$. MDM diffusion model of current task is then finetuned using MDM technique using Real current task Data $\mathcal{D}_{i}^{\text{real}}$ and randomly sampled small batch of $\mathcal{D}_{0:i}^{\text{syn}}$. $J_{0:i}$ is subsequently used to obtain $\mathcal{D}_{i}^{\text{aug}}$ by SSIA. The model trains with MDA on the combined dataset.  } 
\label{fig:architecture}
\end{figure}
\subsubsection{Finetuning Multi-Distribution Matching Diffusion Model with LoRA.}
In the $i^{th}$ incremental task, the real data of the current task  $\mathcal{D}_{i}^{\text{real}}$ and the synthetic data of the previous tasks $\mathcal{D}_{0:i}^{\text{syn}}$ (notation 0:i means integers from 0 up to but not including i) generated by fine-tuned diffusion models $J_{0:i}$ is available.  
We use $\mathcal{D}_{i}^{\text{real}}$ and randomly sampled a small batch of $\mathcal{D}_{0:i}^{\text{syn}}$ to fine-tune a multi-distribution matching diffusion model $J_{i}$ using LoRA. The finetuned diffusion model $J_{i}$ can be used to generate synthetic data. Based on LoRA, the cost to finetune and store diffusion models is relatively small.  

\subsubsection{Forming Training Dataset for Current Task.}  
The training dataset $\mathcal{D}_{i}^{\text{train}}$ for the current task consists of three parts, (1) the synthetic data of the previous tasks $\mathcal{D}_{0:i}^{\text{syn}}$ synthesized by fine-tuned diffusion models $J_{0:i}$, (2) the real data of the current task $\mathcal{D}_{i}^{\text{real}}$, and (3) the image augmentation data $\mathcal{D}_{i}^{\text{aug}}$ generated from $J_{i}$. For   $i=0$, the synthetic data are ignored.
The model can then start training by randomly sampling training batches $(x^{\text{train}}, y^{\text{train}})$ from the newly-formed training dataset. 

\subsubsection{Training with Multi-Domain Adaptation.}
For each batch of training data, we adopt the training method with multi-domain adaptation. Specifically, after feature extraction with a CNN backbone  
defined as $F_{i}: \mathbb{R}^{h \times w \times 3} \to \mathbb{R}^d$,  the extracted features go through two branches: a linear classifier  $G_{i}: \mathbb{R}^d \to \mathbb{R}^c$, and a gradient reverse layer (GRL) followed by a linear classifier   $K_{i}: \mathbb{R}^d \to \mathbb{R}^2$.  
During training, $G_{i}$ learns to classify representations of new classes in new tasks without forgetting previous classes, while $K_{i}$ acquires the knowledge of boundaries between diffusion-generated synthetic data and real data.

The details and advantages of the three stages in each incremental learning phase are specified below.

\subsection{Finetuning Multi-Distribution Matching Diffusion Model with LoRA}
\label{sec:mdm}
Previous exemplar-free CIL works either use or alter the sampling pre-trained diffusion models to synthesize data of previous tasks~\cite{sddr, ddgr}.
However, these methods fail to generate realistic data with evident domain gaps (or distribution discrepancies) for the classes in the same incremental task or 
keep consistent generation quality across different incremental tasks. These bottlenecks affect the model's robustness in stability as shown previously in \cref{sec:domaingaps}.

\subsubsection{Multi-Distribution Matching.}
To address this significant limitation in exemplar-free CIL, inspired by the recent work on training data synthesis~\cite{realfake} with an additional synthetic-to-real distribution-matching technique to enclose the gap between synthetic and real data distributions, we propose a multi-distribution matching (MDM) technique to fine-tune the diffusion model that best fit our exemplar-free CIL setting. 
In specific, when finetuning a diffusion model, we not only match the distributions of the synthetic and real data for the current task but also align the distributions of synthetic data in the current task with that in all previous tasks.
With MDM, the diffusion models can be finetuned by optimizing the following loss:
\begin{equation}
\begin{split}
    \mathcal{L}_{MDM} = & \, || \frac{1}{|\mathcal{D}_{i}^{\text{real}}|+|Z(\mathcal{D}_{0:i}^{\text{syn}})|} \sum_{j=1}^{|\mathcal{D}_{i}^{\text{real}}|+|Z(\mathcal{D}_{0:i}^{\text{syn}})|} \left( \epsilon - \epsilon_{\theta}(x_t^\prime, t) \right) ||_\mathcal{H}^2 \\
    & \leq \frac{1}{|\mathcal{D}_{i}^{\text{real}}|} \sum_{j=1}^{|\mathcal{D}_{i}^{\text{real}}|} || \epsilon - \epsilon_{\theta}(x_t, t) ||_\mathcal{H}^2 = \mathcal{L}_{diff}.
\end{split}
\end{equation}
where $x^\prime \in R_{i}+ Z(S_{0:i})$ and $x \in R_{i}$. Here $Z$ is a random selection function to incorporate only a small portion of synthetic data of past tasks $S_{0:i}$ for multi-distribution matching purposes. $\epsilon_{\theta}$ is the noise predictor for latent space noised $x_t$ with noise $\epsilon$. And $\mathcal{H}$ denotes it's in the universal Reproducing Kernel Hilbert
Space. The Loss is further constraint by the original stable diffusion loss on only $\mathcal{D}_{i}^{\text{real}}$ to emphasize while MDM is focused on multi-distribution matching crossing all training phase data, it should not compromise the fundamental denoising or data generation ability of the model of current real task classes.
We also provide detailed deduction and proof for this equation in the Appendix.

In this way, the synthetic images generated using the diffusion models with the proposed MDM are of uniform quality in different classes and tasks. More importantly, the distribution discrepancies or demain gaps between synthetic and real images become smaller, which fundamentally alleviates the potential domain bias problems and achieves better CIL performance. 

\subsection{Forming Current Task Training Dataset}
Synthetic Data Augmentation has proven to enhance the model performance on various computer vision tasks due to its ability to enlarge training data distribution~\cite{synaug, synaug2}. In exemplar-free CIL, various image augmentation techniques~\cite{sddr, pass, il2a} are frequently adopted. 
Therefore, when structuring the current task training dataset, aside from synthetic previous-task data generated by diffusion models $J_{0:i}$, and the real data of the current task, we further incorporate data augmentation $\mathcal{D}_{i}^{\text{aug}}$ with synthetic data of current task from $J_{i}$. 
However, in enhancing and aligning our method, we propose a different data augmentation technique, \ie selective synthetic image augmentation (SSIA), to obtain $\mathcal{D}_{i}^{\text{aug}}$. In specific, rather than finetuning and utilizing generative models after each training phase~\cite{abd, rdfcil, ddgr}, at the beginning phase of each task $i$, we finetune a MDM diffusion model $J_{i}$ using LoRA as proposed in \cref{sec:mdm}.

We generate twice the number of synthetic data as real data for the current task $i$ and filter out the same number (or less) of distributional representative synthetic images as real data. It includes the following key steps.
\begin{itemize}
\item Calculate each generated class mean and create covariance matrices.
\begin{equation}
\mu_{cn}^{\text{gen}} = \frac{1}{|\mathcal{D}_{cn}^{\text{gen}}|} \sum_{\mathbf{x} \in \mathcal{D}_{cn}^{\text{gen}}} \mathbf{x}, \text{where } cn \in \mathcal{CN},
\end{equation}
\begin{equation}
\text{Cov}_{cn}^{\text{gen}} = \frac{1}{|\mathcal{D}_{cn}^{\text{gen}}|-1} \sum_{\mathbf{x} \in \mathcal{D}_{cn}^{\text{gen}}} (\mathbf{x} - \mu_{cn}^{\text{gen}})(\mathbf{x} - \mu_{cn}^{\text{gen}})^T,
\end{equation}
where $\mathcal{CN}$ denotes all classes in the current task.

\item Sample the generated images for each current task class
\begin{equation}
\mathbf{x}_{i}^{cn} \sim \mathcal{N}(\mu_{cn}^{\text{gen}}, \text{Cov}_{cn}^{\text{gen}}),
\end{equation}

\item Calculate a selected threshold for synthetic image selection and construct the image augmentation dataset.
\begin{equation}
\tau_{cn}^{\text{gen}} = k \cdot \sqrt{\text{diag}(\text{Cov}_{cn}^{\text{gen}})},
\end{equation}
\begin{equation}
\mathcal{D}_i^{\text{aug}} = \bigcup_{cn=1}^{\mathcal{CN}} \{ \mathbf{x}_{i}^{cn} \,|\, \|{\mathbf{x}_{i}^{cn} - \mu_{cn}^{\text{gen}}\| \leq \tau_{cn}^{\text{gen}}} \}.
\end{equation}
\end{itemize}

With SSIA, our method can benefit for multiple reasons.  MDM  mitigates the domain gaps between synthetic data in different tasks and the diffusion models can generate more realistic high-quality images for SSIA.  
This helps to enhance the model's stability since domain-aligned training data can contribute to preventing feature embedding domain bias problems in exemplar-free CIL settings.
SSIA can enable the model to better build knowledge for new classes. The model is capable of learning from the classes of current task trained with broader data distributions. The quality of images in SSIA is strong and representative since the synthetic images are selected from clusters around the class mean and span a calculated range with a broader class distribution.
Moreover, the current task training dataset consists of both real and synthetic domains, which fortifies the multi-domain adaptation capabilities in our framework later discussed in  \cref{sec:mda}.

\subsection{Training with Multi-Domain Adaptation}
\label{sec:mda}
Even with the multi-distribution matching technique, we still notice a nontrivial domain gap between synthetic data and real data in the training dataset. This domain gap will inevitably affect the model performance on classifying previous-task images during incremental learning, as shown in \cref{sec:domaingaps}.
Previous exemplar-free CIL works mainly adopt knowledge distillation techniques~\cite{abd, rdfcil} to implicitly avoid the model favoring domains over classes, \ie aiming to enable the model to classify whether it is real or synthetic rather than its true labels.
However, knowledge distillation still fails to address the domain gap problem with low classification performance in CIL and a high computation complexity.

\subsubsection{Multi-Domain Adaptation.} 
To deal with these problems, we propose to reformulate exemplar-free CIL as a task-agnostic multi-domain adaption problem. Inspired by domain-adversarial training~\cite{dat}, for each task $\mathcal{T}_{i}$, after the original CNN backbone, besides the original linear classifier $G_{i}: \mathbb{R}^d \to \mathbb{R}^c$ for class label classification,   we further construct an additional branch with a gradient reverse layer followed by another linear classifier $K_{i}: \mathbb{R}^d \to \mathbb{R}^2$ for domain prediction.

Hence we can formulate our exemplar-free CIL training approach in each task $\mathcal{T}_{i}$ as optimizing the following:
\begin{equation}
\mathcal{L}_{i}^{\text{train}} = \mathcal{L}_{i}^{\text{cls}} + \mathcal{L}_{i}^{\text{da}}.
\end{equation}
where
\begin{equation}
\mathcal{L}_{i}^{\text{cls}} = -\frac{1}{|\mathcal{D}_{i}^{\text{train}}|} \sum_{\mathbf{x} \in \mathcal{D}_{i}^{\text{train}}} y_{c} \log \left( G_i \left( F_i\left(\mathbf{x} \right) \right) \right),
\end{equation}
and
\begin{equation}
\mathcal{L}_{i}^{\text{da}} = -\frac{1}{|\mathcal{D}_{i}^{\text{train}}|} \sum_{\mathbf{x} \in \mathcal{D}_{i}^{\text{train}}} \left[ y_{d} \log \left( K_i \left( F_i\left( \mathbf{x} \right) \right) \right) + (1 - y_{d}) \log \left( 1 - \left( K_i \left( F_i\left( \mathbf{x} \right) \right) \right) \right) \right].
\end{equation}
Here $y_{c}$ represents the ground truth label for class $c$, and $y_{d}$ represents the ground truth domain label $d$.
The model needs to not only learn to classify the image but also distinguish whether it is real or synthetic.

Different from traditional domain-adversarial training with a focus on single target domain (real) data only, in our exemplar-free CIL setting, our model benefits from training both classification and domain branches using both target (real) and source (synthetic) domain data in each incremental task $\mathcal{T}_{i}$. 
For learning classification knowledge in  $\mathcal{T}_{i}$, synthetic data is a crucial key for reviewing previous knowledge while real data contributes to gaining new knowledge. 
For learning multi-domain adaptation knowledge, adopting a mixture of data from both domains can contribute to differentiating and adapting to the distinct characteristics of each domain.

By reforming exemplar-free CIL as a straightforward task-agnostic multi-domain adaption problem, our method enjoys the following advantages. (i) Our model framework keeps simple without any cumbersome parts, which benefits incremental training efficiency.
(ii) More importantly, our model is robust in both stability and plasticity since it is fully capable of learning important feature knowledge from both label classification and domain classification (synthetic \vs real) in each task. 
(iii) Our proposed method can not only perform well on a test dataset consisting of entirely real data but also elaborate to perform well on entirely synthetic test data and combined image groups (see Appendix)
, which better simulates the continual learning scenarios in real-world settings.
\section{Experiment}
\label{sec:experiment}
\subsection{Datasets and Evaluation Protocol}
\noindent\textbf{Datasets.} To accurately and fairly evaluate our method in comparison with baselines, we use two representative datasets CIFAR100~\cite{cifar} and ImageNet100\\~\cite{ucir}, which are widely adopted in CIL. 
CIFAR100 consists of 100 classes, each containing 500 training and 100 test images with the resolution 32$\times$32$\times$3. ImageNet100 is a randomly sampled subset of ImageNet1000~\cite{imagenet}, consisting of 100 classes each with 1300 training and 50 test images of various sizes.

{\setlength{\tabcolsep}{5.7pt}
\renewcommand\arraystretch{1.2}
\begin{table}[tb]
  \centering
  \caption{Evaluation results on CIFAR100 with protocol that equally split 100 classes into $N$ tasks. The best results are in bold.}
  \label{tab:cifar100}
  \begin{tabular}{l c c| c c |c c}
    \hline
    \multirow{2}{*}{Approach} & \multicolumn{2}{c}{$N=5$} & \multicolumn{2}{c}{$N=10$}  & \multicolumn{2}{c}{$N=20$} \\
    \cline{2-7}
    & ${Acc}_{avg}$ & $Acc_L$ & ${Acc}_{avg}$ & $Acc_L$ & ${Acc}_{avg}$ & $Acc_L$ \\
    \hline
    Upper Bound & & 70.67 & & 70.67 & & 70.67 \\ 
    \cline{2-7}
    ABD~\cite{abd} (ICCV 2021) & 60.78 & 44.74 & 54.00 & 34.48 & 43.32 & 21.18 \\
    PASS~\cite{pass} (CVPR 2021) & 63.31 & 49.11 & 52.01 & 36.08 & 41.84 & 27.45 \\
    IL2A~\cite{il2a} (NeurIPS 2021) & 58.67 & 45.34 & 43.28 & 24.49 & 40.54 & 21.15 \\
    R-DFCIL~\cite{rdfcil} (ECCV 2022) &  64.67 & 50.24 & 59.18 & 42.17 & 49.74 & 31.46 \\
    SSRE~\cite{ssre} (CVPR 2022) & 56.96 & 43.05 & 43.41 & 29.25 & 31.07 & 16.99 \\
    FeTril~\cite{fetril} (WACV 2023) & 58.68 & 42.67 & 47.14 & 30.28 & 37.25 & 20.62 \\
    SEED~\cite{seed} (ICLR 2024) & 63.05 & 52.14 & 62.04 & 51.42 & 57.42 &  42.87\\
    \cline{2-7}
    \textbf{Ours} & \textbf{69.77} & \textbf{62.21} & \textbf{68.05} & \textbf{58.40} & \textbf{67.10} & \textbf{57.11}\\
    \hline
  \end{tabular}
\end{table}
}

\noindent\textbf{Incremental Settings.} Following prior works~\cite{abd,rdfcil, seed}, for CIFAR100 and ImageNet100 datasets, we split the classes equally into $N=5$, 10, or 20 tasks (\eg, each task has 5 classes if $N=20$). For all approaches, we use the same random seed to randomly shuffle class orders for all datasets.
Following previous works~\cite{abd, pass, il2a, rdfcil, ssre, fetril, seed}, the classification accuracy is defined as
\begin{equation}
Acc_i = \frac{1}{\lvert \mathcal{D}_{0:i+1}^{test} \rvert}
\sum_{(\mathbf{x}, y) \in \mathcal{D}_{0:i+1}^{test}} 
\mathds{1} \left(\hat{y} = y\right), 
~\text{where}~
\hat{y} = 
\mathop{\arg \max}_{j \in \mathcal{C}_{i}}
G_{i}^{(j)}(F_i(\mathbf{x})).
\end{equation}
We report both the final accuracy from the last task $Acc_{L}$ and the average incremental accuracy 
averaged over all incremental tasks ${Acc}_{avg} = \frac{1}{N} \sum_{i=0}^{N-1} Acc_i$.
\noindent\textbf{Implementation Details.}
For a fair comparison, for CIFAR100, following previous works~\cite{abd,rdfcil}, we use a modified 32-layer ResNet~\cite{resnet} as the backbone for all approaches. For our model, we train with SGD optimizer for 120 epochs. The learning rate is initially set to 0.1 with a decay factor of 0.1 after 100 epochs. The weight decay is set to 0.0002 and batch size of 128. For ImageNet100, we use ResNet18~\cite{resnet} as the backbone for all methods. For our training, the SGD optimizer is adopted to train  40 epochs. The learning rate is initially set to 0.1 with a decay factor of 0.1 after 30 epochs. The weight decay is set to 0.0001 and batch size of 128. We train and report all methods from scratch with original implementations.

\begin{figure}[tb]
  \centering
  \begin{subfigure}{\linewidth}
    \centering
    \begin{subfigure}{0.32\linewidth}
      \includegraphics[width=\linewidth]{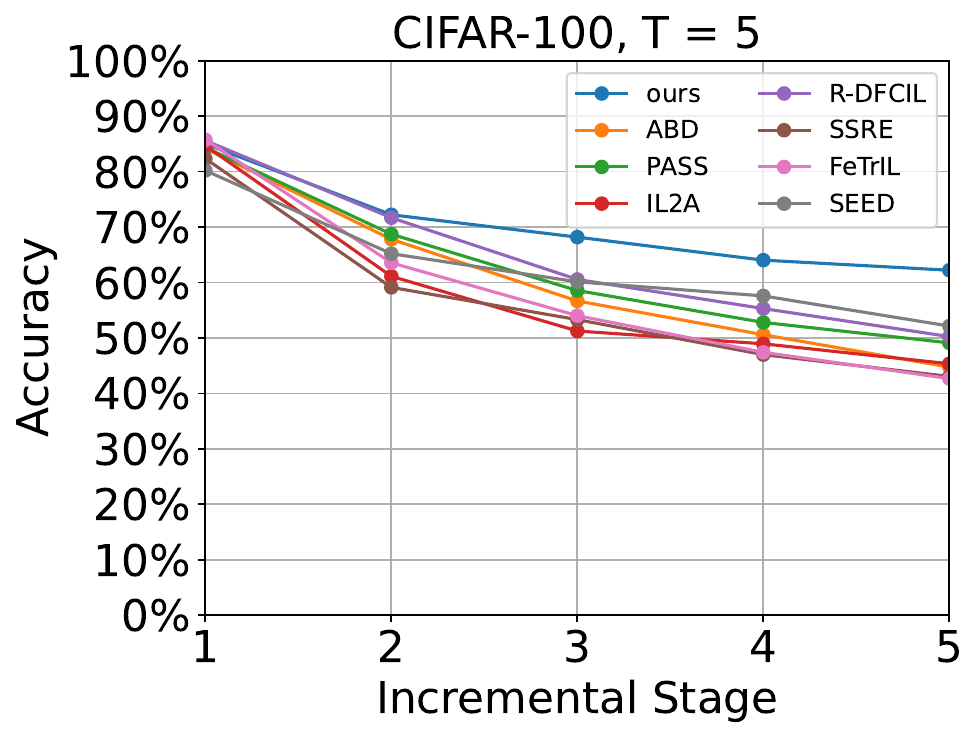}
      \caption{5 tasks, 20 classes/task}
    \end{subfigure}
    \begin{subfigure}{0.32\linewidth}
      \includegraphics[width=\linewidth]{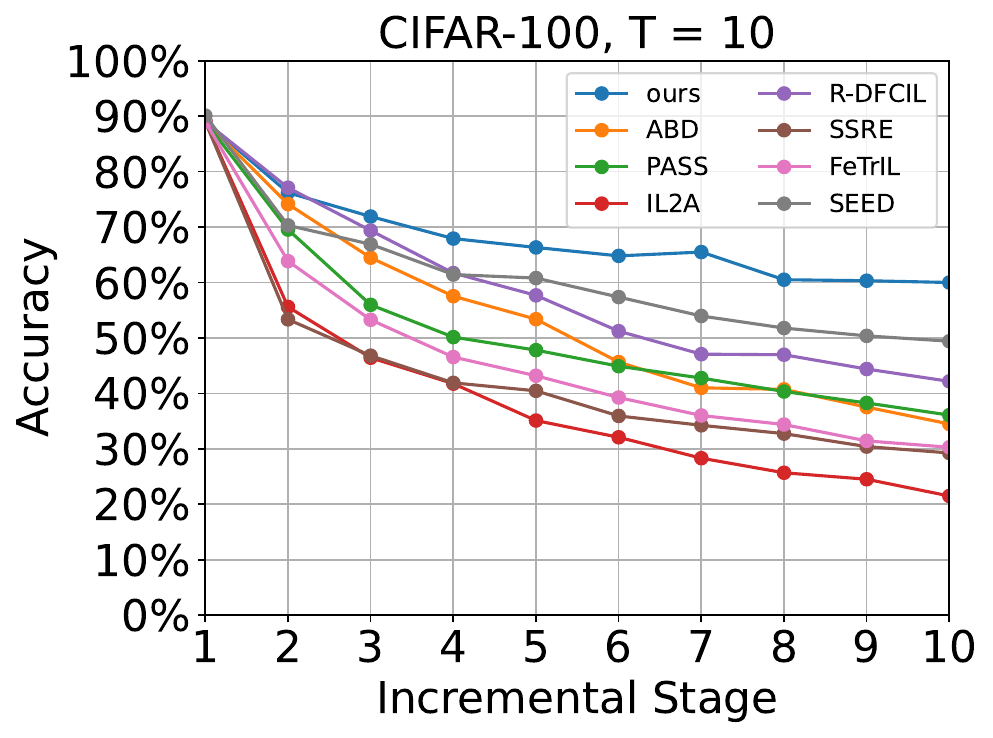}
      \caption{10 tasks, 10 classes/task}
    \end{subfigure}
    \begin{subfigure}{0.32\linewidth}
      \includegraphics[width=\linewidth]{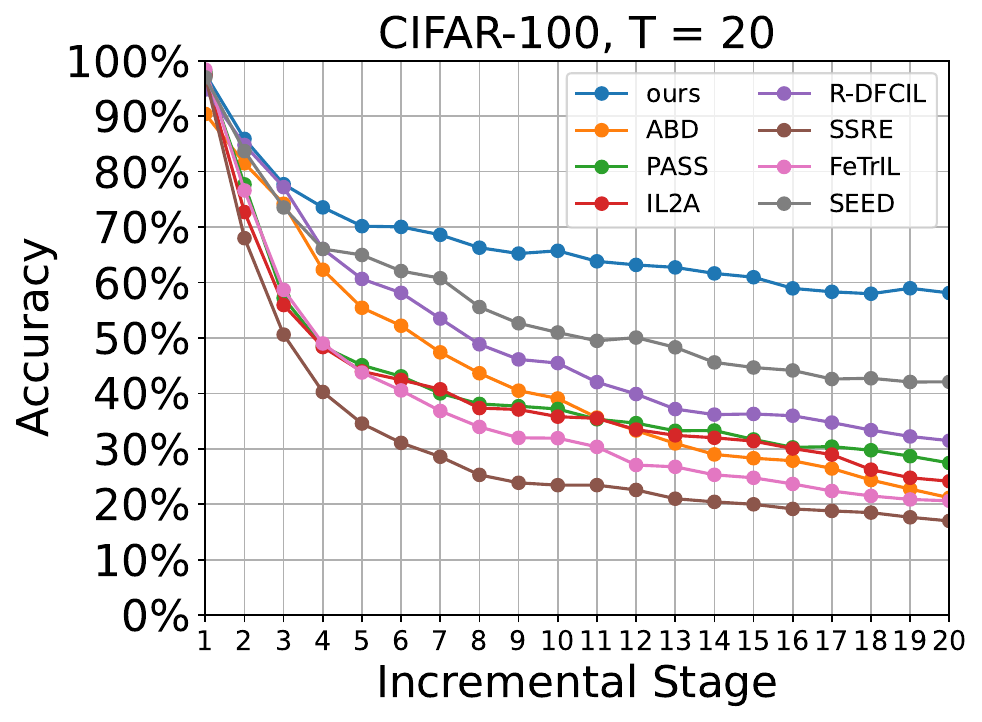}
      \caption{20 tasks, 5 classes/task}
    \end{subfigure}
  \end{subfigure}
  \caption{\textbf{Classification Accuracy of Each Incremental Task on CIFAR100}. Our method greatly outperforms all  data-free CIL baselines in all incremental settings.
  }
  \label{fig:cifar100}
\end{figure}

\subsection{Results and Analysis}
\subsubsection{CIFAR100.}
We report the results of our method and  SOTA exemplar-free CIL methods on CIFAR100 in \cref{tab:cifar100}. As observed, our method achieves the highest average and final accuracy among all approaches with non-marginal improvements.  Moreover, as CIL becomes more difficult with a larger $N$ (such as 20), the baselines suffer from significant accuracy drop (such as from 51.42\%  to 42.87\% for SEED~\cite{seed} when increasing $N$ from 10 to 20), while our method still maintains high accuracy close to that of smaller $N$ (such as our final accuracy from 58.4\% to  57.11\%) with larger improvements over baselines. Notably,  compared with SOTA exemplar-free CIL method  SEED(ICLR 2024)~\cite{seed}, when $N=20$, our method is 9.68 percent more accurate for the average incremental accuracy ${Acc}_{avg}$ and  14.24 percent more accurate for the final accuracy ${Acc}_{L}$.

{\setlength{\tabcolsep}{5.7pt}
\renewcommand\arraystretch{1.2}
\begin{table}[tb]
  \centering
  \caption{Evaluation on ImageNet100 with protocol that equally split 100 classes into $N$ tasks.}
  \label{tab:imagenet100}
  \begin{tabular}{l c c| c c |c c}
    \hline
    \multirow{2}{*}{Approach} & \multicolumn{2}{c}{$N=5$} & \multicolumn{2}{c}{$N=10$}  & \multicolumn{2}{c}{$N=20$} \\
    \cline{2-7}
    & ${Acc}_{avg}$ & $Acc_L$ & ${Acc}_{avg}$ & $Acc_L$ & ${Acc}_{avg}$ & $Acc_L$ \\
    \hline
    Upper Bound & & 80.41 & & 80.41 & & 80.41 \\ 
    \cline{2-7}
    ABD~\cite{abd} (ICCV 2021) & 67.12 & 52.00 & 57.06 & 35.66 & 45.75 & 22.10 \\
    PASS~\cite{pass} (CVPR 2021) & 55.75 & 39.50 & 33.75 & 16.18 & 27.30 & 18.24 \\
    IL2A~\cite{il2a} (NeurIPS 2021) & 62.66 & 48.91 & 43.46 & 26.04 & 35.59 & 20.72 \\
    R-DFCIL~\cite{rdfcil} (ECCV 2022) & 68.42 & 53.50 & 59.36 & 42.70 & 49.99 & 30.80 \\
    SSRE~\cite{ssre} (CVPR 2022) & 52.25 & 37.76 & 46.00 & 29.28 & 34.96 & 18.90 \\
    FeTril~\cite{fetril} (WACV 2023) & 58.40 & 41.44 & 46.44 & 27.92 & 37.64 & 20.62 \\
    SEED~\cite{seed} (ICLR 2024) & 69.08 & 58.17 & 67.55 & 55.17 & 62.26 & 45.77 \\
    \cline{2-7}
    \textbf{Ours} & \textbf{74.85} & \textbf{67.26} & \textbf{73.87} & \textbf{67.02} & \textbf{72.51} & \textbf{68.68}\\
    \hline
  \end{tabular}
\end{table}
}

We further present the detailed incremental accuracy of various learning phases for  $N=5$, 10, and 20 on CIFAR100 in \cref{fig:cifar100}. We observe that our curve drops significantly slower than all baseline methods with the highest accuracy at various phases, demonstrating our superior performance to mitigate the forgetting of previously learned knowledge over baseline methods.
\begin{figure}[tb]
  \centering
  \begin{subfigure}{\linewidth}
    \centering
    \begin{subfigure}{0.32\linewidth}
      \includegraphics[width=\linewidth]{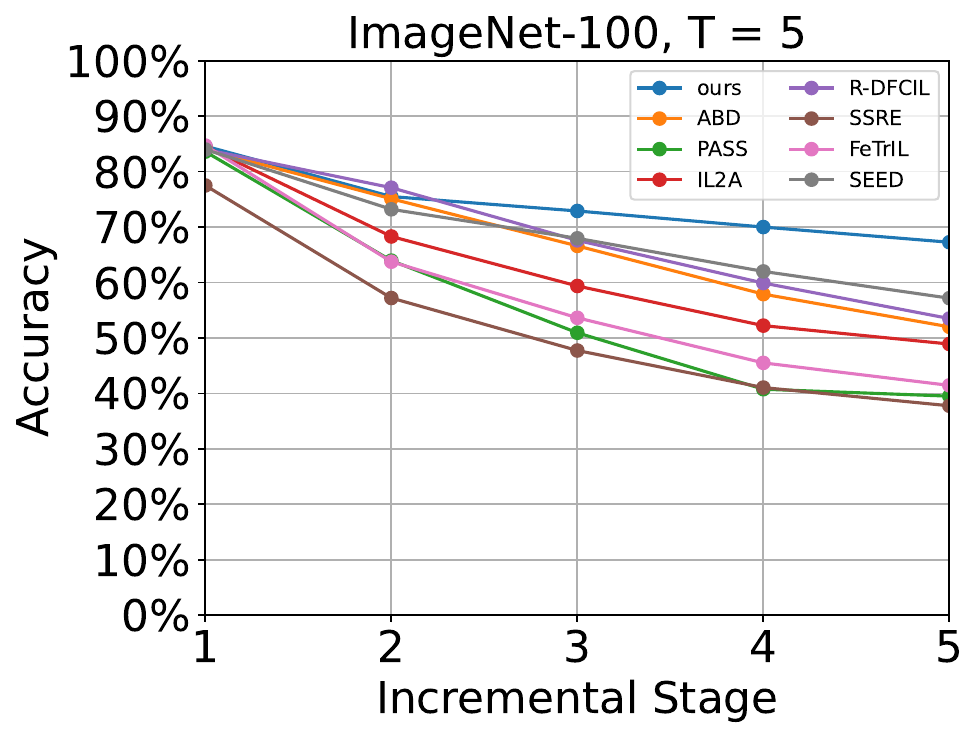}
      \caption{5 tasks, 20 classes/task}
    \end{subfigure}
    \begin{subfigure}{0.32\linewidth}
      \includegraphics[width=\linewidth]{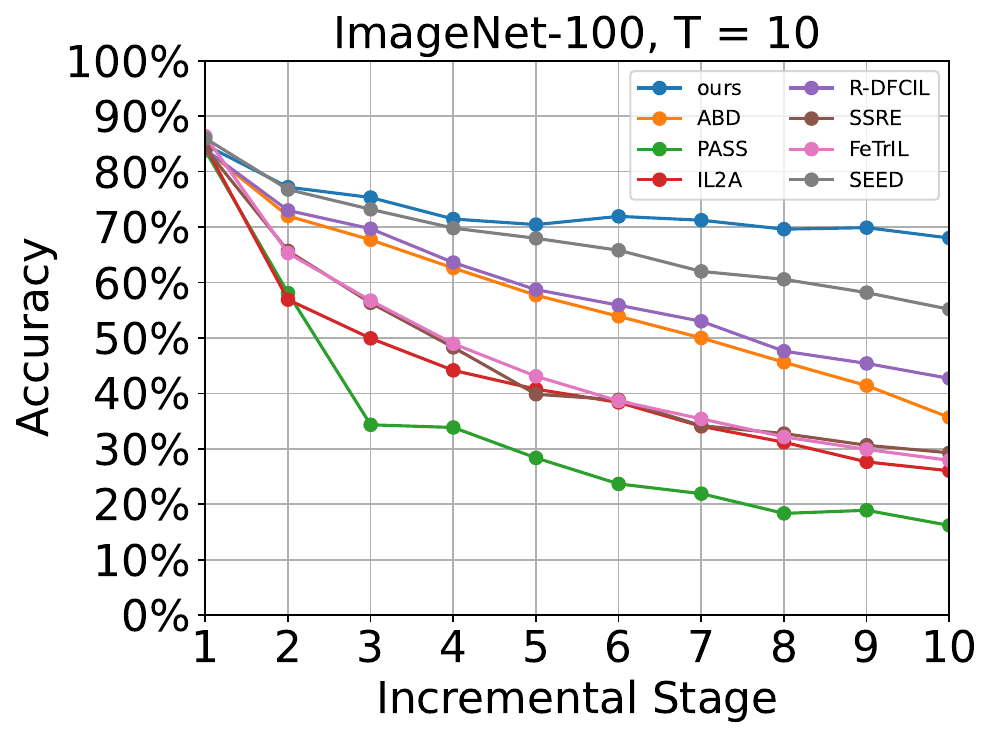}
      \caption{10 tasks, 10 classes/task}
    \end{subfigure}
    \begin{subfigure}{0.32\linewidth}
      \includegraphics[width=\linewidth]{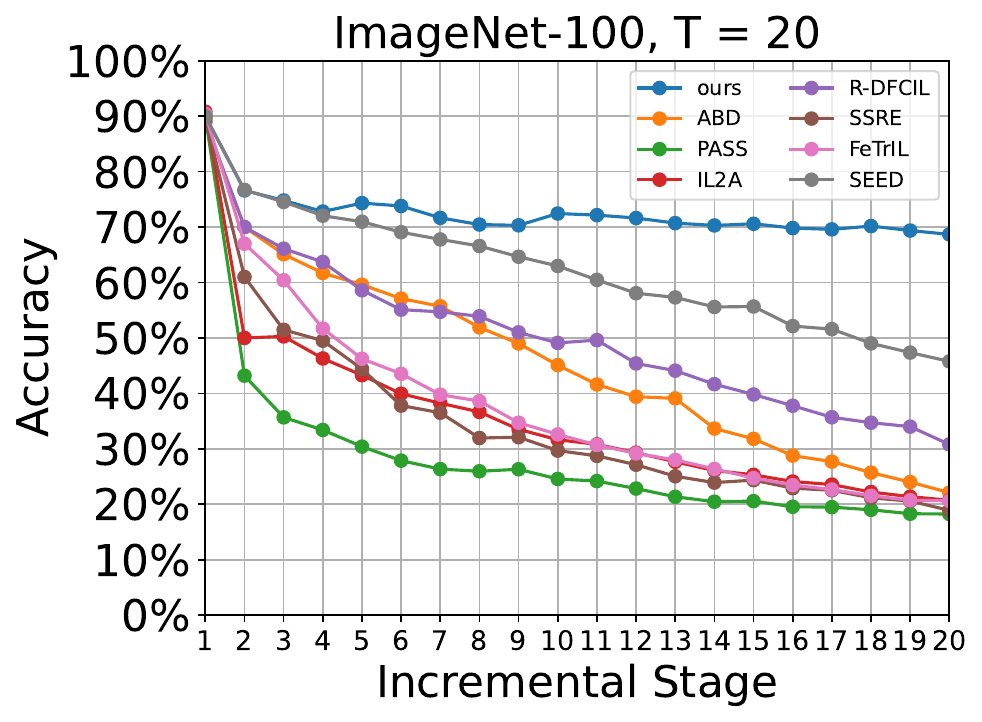}
      \caption{20 tasks, 20 classes/task}
    \end{subfigure}
  \end{subfigure}
  \caption{\textbf{Incremental Accuracy on ImageNet100}. Our method greatly outperforms all baseline methods in all incremental settings. Our method achieves  more significant improvements in more incremental task settings (\eg increase $N$ from 5 to 10 or to 20)
  }
  \label{fig:imagenet100}
\end{figure}

\subsubsection{ImageNet100.}
In \cref{tab:imagenet100}, we present the results of our method and SOTA exemplar-free CIL methods on ImageNet100. Similarly, our method outperforms all baselines in terms of the average accuracy and final accuracy with non-marginal improvements. As CIL becomes more difficult with a larger $N$, the advantages or improvements of our method become more significant. 
Compared with  SOTA exemplar-free CIL method seed~\cite{seed}, for $N=20$, our method is  10.25 percent more accurate for  ${Acc}_{avg}$ and 22.91 percent more accurate for  ${Acc}_{N}$. 

The detailed incremental accuracy of various learning phases for  $N=5$, 10, and 20 on ImageNet100 are presented in \cref{fig:imagenet100}. As observed, our method keeps the highest accuracy at almost all of the learning phases or stages. As it goes through more learning phases,  our method can maintain almost consistent accuracy, outperforming baselines (which suffer from significant accuracy drops) with larger improvements.  The results demonstrate that our method performs much better to mitigate the catastrophic forgetting problem in CIL. 

\subsection{Ablation Studies}
We ablate the three major components in our method on ImageNet100  with $N=5$.
In each phase, $5$ new classes are learned. We present our ablation results in \cref{tab:ablation}.
The results demonstrate that all proposed components contribute greatly. We further show that all three components are crucial to achieving better plasticity \vs stability balance through an ablation study
in \cref{fig:ablation}.

{\setlength{\tabcolsep}{14.4pt}
\renewcommand\arraystretch{1.25}
\begin{table}[tb]
  \centering
  \caption{Abalation Study results of comparison between our method with all components and without the multi-distribution-matching diffusion model (MDM), without multi-domain adaptation reformation (MDA), and without selective synthetic image augmentation (SSIA). The ablation study is conducted on ImageNet100 with $N=5$. }
  \label{tab:ablation}
  \begin{tabular}{lcccccc|c}
  \hline
   MDM && MDA && SSIA && $Acc_{avg}$ & $Acc_{N}$ \\
   \hline
   \ding{55} && \ding{51} &&  \ding{51} && 59.71 & 51.17 \\
   \cline{7-8}
   \ding{51} && \ding{55} &&  \ding{51} && 65.29 & 55.22\\
   \cline{7-8}
   \ding{51} && \ding{51} &&  \ding{55} && 62.37 & 52.94 \\
   \cline{7-8}
   \ding{51} && \ding{51} &&  \ding{51} && 74.85 & 67.26 \\
  \hline
  \end{tabular}
\end{table}
}

\subsubsection{Multi-Distribution Matching(MDM).}
Without finetuning diffusion models with a multi-distribution matching technique, the average accuracy $Acc_{avg}$ drops by 15.14 percent (74.85\% \vs 59.71\%), and the final classification accuracy $Acc_{N}$ drops by 16.09 percent (67.26\% \vs 51.17\%). From \cref{fig:ablation}, we also observe that MDM serves a crucial role in reviewing previous knowledges (\ie stability).

\subsubsection{Multi-Domain Adaptation (MDA).}
Without reforming exemplar-free CIL into a multi-domain adaptation problem, the average accuracy $Acc_{avg}$ drops by 9.56 percent, and the final accuracy $Acc_{N}$ drops by 12.04 percent. MDA also contributes to building model stability as shown in \cref{fig:ablation}.

\subsubsection{Selective Synthetic Image Augmentation (SSIA).}
\begin{figure}[tb]
\centering
\includegraphics[width=270pt]{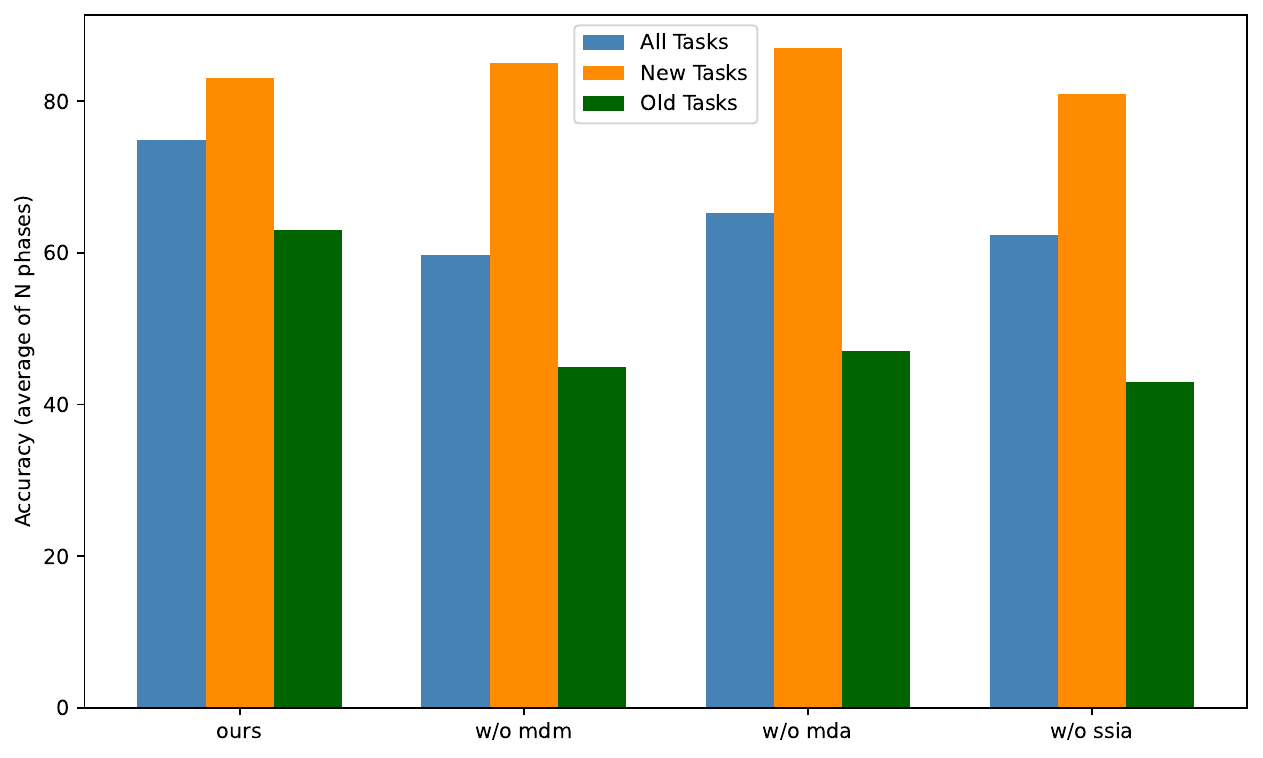}
\caption{\textbf{Ablation Study about Stability-Plasticity Balance}. Our method with all three components shows a better balance \vs w/o each of the three components.}
\label{fig:ablation}
\end{figure}

Without further enhancement from selective synthetic image augmentation, the average accuracy $Acc_{avg}$ drops by 12.48 percent, and the final accuracy $Acc_{L}$ drops by 14.97 percent. Furthermore, \cref{fig:ablation} shows that SSIA helps the model not only learn new knowledge (\ie plasticity) but also remember the knowledge from previous tasks.

\section{Conclusion}
\label{sec:conclusion}
In this paper, we introduce a novel exemplar-free CIL approach to address catastrophic forgetting and stability and 
plasticity imbalance caused by the domain gap between synthetic and real data. Specifically, our method generates synthetic data using multi-distribution matching (MDM) diffusion models to explicitly bridge the domain gap and unify quality among all training data. Selective synthetic image augmentation (SSIA) is also applied to enlarge training data distribution, enhancing the model's plasticity and bolstering the efficacy of our method's final component, multi-domain adaptation (MDA). With the proposed integrations, our method then reforms exemplar-free CIL to a multi-domain adaptation problem to implicitly address the domain gap problem during incremental training.
Our method achieves state-of-the-art performance in various exemplar-free CIL settings on CIFAR100 and ImageNet100 benchmarks. In the ablation study, we
proved that each component of our method is significant to best perform in exemplar-free CIL.
\subsubsection{Limitations and Future Works}
One potential limitation of our method is the training time for each incremental task. In specific, the time to finetune a generative model using LoRA. This limitation is very common in exemplar-free methods that utilize synthetic data. In our case, we deduce in each incremental phase, the time to finetune an MDM diffusion model is proportional to the number of new classes to learn. In future work, we aim to explore strategies to streamline this process, thereby enhancing a shorter exemplar-free CIL training process.

\section*{Acknowledgement}
This work is partially supported by the Army Research Office/Army Research Laboratory via grant W911-NF-20-1-0167 to Northeastern University, National Science Foundation CCF-1937500.
\bibliographystyle{plain}
\bibliography{main}
\end{document}